# Pattern Matching and Discourse Processing in Information Extraction from Japanese Text

**Tsuyoshi Kitani**                                             TKITANI@CS.CMU.EDU
*Center for Machine Translation*
*Carnegie Mellon University*
*Pittsburgh, PA 15213 USA*

**Yoshio Eriguchi**                                             ERIGUCHI@RD.NTTDATA.JP
**Masami Hara**                                                 MASAMI@RD.NTTDATA.JP
*Development Headquarters*
*NTT Data Communications Systems Corp.*
*66-2 Horikawa-cho, Saiwai-ku, Kawasaki-shi, Kanagawa 210 JAPAN*

## Abstract

Information extraction is the task of automatically picking up information of interest from an unconstrained text. Information of interest is usually extracted in two steps. First, sentence level processing locates relevant pieces of information scattered throughout the text; second, discourse processing merges coreferential information to generate the output. In the first step, pieces of information are locally identified without recognizing any relationships among them. A key word search or simple pattern search can achieve this purpose. The second step requires deeper knowledge in order to understand relationships among separately identified pieces of information. Previous information extraction systems focused on the first step, partly because they were not required to link up each piece of information with other pieces. To link the extracted pieces of information and map them onto a structured output format, complex discourse processing is essential. This paper reports on a Japanese information extraction system that merges information using a pattern matcher and discourse processor. Evaluation results show a high level of system performance which approaches human performance.

## 1. Introduction

In recent information extraction systems, most individual pieces of information to be extracted directly from a text are usually identified by key word search or simple pattern search in the preprocessing stage (Lehnert et al., 1993; Weischedel et al., 1993; Cowie et al., 1993; Jacobs et al., 1993). Among the systems presented at the Fifth Message Understanding Conference (MUC-5), however, the main architectures ranged from pattern matching to full or fragment parsing (Onyshkevych, 1993). Full or fragment parsing systems, in which several knowledge sources such as syntax, semantics, and domain knowledge are combined at run-time, are generally so complicated that changing a part of the system tends to affect the other components. In past information extraction research, this interference has slowed development (Jacobs, 1993; Hobbs et al., 1992). A pattern matcher, which identifies only patterns of interest, is more appropriate for information extraction from texts in narrow domains, since this task does not require full understanding of the text.

TEXTRACT, the information extraction system described here, uses a pattern matcher similar to SRI's FASTUS pattern matcher (Hobbs et al., 1992). The matcher is implemented as a finite-state automaton. Unlike other pattern matchers, TEXTRACT's matcher deals with word matching problems caused by the word segmentation ambiguities often found in Japanese compound words.

The goal of the pattern matcher is to identify the *concepts* represented by words and phrases in the text. The pattern matcher first performs a simple key-word-based *concept search*, locating individual words associated with concepts. The second step is a *template pattern search* which locates phrasal patterns involving critical pieces of information identified by the preprocessor. The template pattern search identifies relationships between matched objects in the defined pattern as well as recognizing the concept behind the relationship. One typical concept is the relationship of "economic activity" which companies can participate in with each other.

It is usually difficult to determine the relationships among pieces of information which have been identified in separate sentences. These relationships are often stated implicitly, and even if the text explicitly mentions them the descriptions are often located far enough apart to make detection difficult. Although the importance of discourse processing for information extraction has been emphasized in the Message Understanding Conferences (Lehnert & Sundheim, 1991; Hirschman, 1992), no system presented has satisfactorily addressed the issue.

The discourse processor in TEXTRACT is able to correlate individual pieces of information throughout the text. TEXTRACT merges concepts which the pattern matcher has identified separately (and usually in different sentences) when the concepts involve the same companies. TEXTRACT can unify multiple references to the same company even when the company name is missing, abbreviated, or pronominalized. Furthermore, the processor segments the discourse to isolate portions of text relevant to a particular conceptual relationship. The discourse segmentation lessens the chance of merging unrelated information (Kitani, 1994).

This paper analyzes some evaluation results for TEXTRACT's discourse module and describes the TIPSTER/MUC-5 evaluation results in order to assess overall system performance.

## 2. TIPSTER information extraction task

The goal of the TIPSTER project sponsored by ARPA is to capture information of interest from English and Japanese newspaper articles about corporate joint ventures and microelectronics. A system must fill a generic template with information extracted from the text by a fully automated process. The template is composed of several objects, each containing several slots. Slots may contain pointers to related objects (Tipster, 1992). Extracted information is to be stored in an object-oriented database.

In the joint ventures domain, the task is to extract information concerning joint venture relationships which organizations form and dissolve. The template structure represents these relationships with tie-up-relationship objects, which contain pointers to organization entity objects representing the organizations involved. Entity objects contain pointers to other objects such as person and facility objects, as shown in Figure 1.

In the microelectronics domain, extraction focuses on layering, lithography, etching, and packaging processes in semiconductor manufacturing for microchip fabrication. The entities

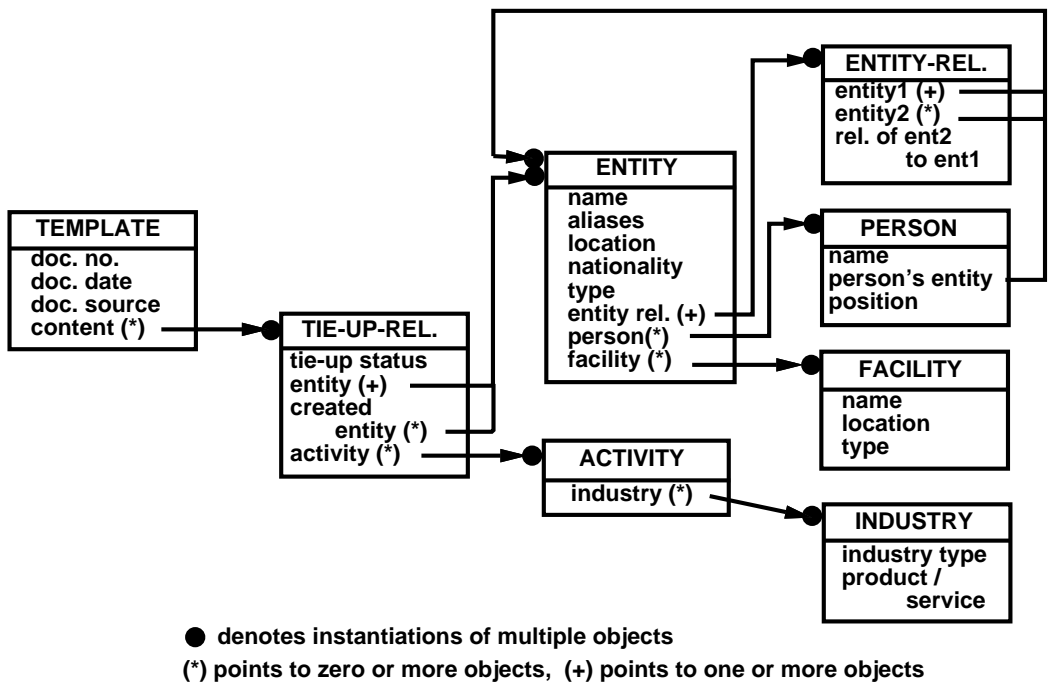

**● denotes instantiations of multiple objects**
**(*) points to zero or more objects, (+) points to one or more objects**

Figure 1: Object-oriented template structure of the joint ventures domain

extracted include manufacturer, distributor, and user, in addition to detailed manufacturing information such as materials used and microchip specifications such as wafer size and device speed. The microelectronics template structure is similar to that of the joint ventures but has fewer objects and slots.

Both of these extraction tasks must identify not only individual entities but also certain relationships among them. Often, however, a particular piece of extracted information describes only part of a relationship. This partial information must be merged with other pieces of information referring to the same entities. For merging to produce correct results, therefore, correct identification of entity references is crucial.

## 3. Problem definition

This section first describes word matching problems caused by the word segmentation ambiguities. Difficulties of reference resolution of company names are then explained. Issues of discourse segmentation and concept merging are also discussed using an example text.

### 3.1 Word segmentation

Japanese word segmentation in the preprocessor gives rise to a subsequent *under-matching* problem. When a key word in the text is not found in the word segmentor's lexicon, the segmentor tends to divide it into separate words. With our current lexicon, for example, the compound noun " 提携解消 " (*teikei-kaisyo*), consisting of two words, " 提携 " (*teikei*: joint venture) and " 解消 " (*kaisyo*: dissolve), is segmented into the two individual nouns. Thus a key word search for " 提携解消 " (*teikei-kaisyo*) does not succeed in the segmented sentence.

On the other hand, the pattern matching process allows, by default, partial matching between a key word and a word in the text. " 提携 " (*teikei*) and " 業務 提携 " (*gyoum-teikei*), both meaning "a joint venture", can both be matched against the single key word " 提携 " (*teikei*). This flexibility creates an *over-matching* problem. For example, the key word " シリコン " (*silicon*) matches " 二酸化 シリコン " (*nisanka-silicon*: silicon dioxide), although they are different materials to be reported in the microelectronics domain. These segmentation difficulties for compound nouns also cause major problems in word-based Japanese information retrieval systems (Fujii & Croft, 1993).

## 3.2 Company name references

In the corporate joint ventures domain, output templates mostly describe relationships among companies (as described in Section 2). Information of interest is therefore found in sentences which mention companies or their activities. It is essential for the extractor to identify topic companies—the main concern of the sentences they appear in—in order to correlate other information identified in the sentence. There are three problems which make it difficult to identify topic companies.

1. Missing subject
   Topic companies are usually the subject of a sentence. Japanese sentences frequently omit subjects, however—even in formal newspaper articles. The VENIEX system which NEC presented at MUC-5 can identify the company implied by a missing subject if there is an explicit reference to it in the immediately preceding sentence (Doi et al., 1993; Muraki et al., 1993). It is not clear whether VENIEX can resolve the missing reference when the explicit reference appears in a sentence further separated from the subjectless sentence.

2. Company name abbreviations
   As is also seen in English, company names are often abbreviated in a Japanese text after their first appearance. A variety of ways to abbreviate company names in Japanese is given in (Karasawa, 1993). The following examples show some typical abbreviations of Japanese company names:

   (a) a partial word
      " メルセデス・ベンツ " → " ベンツ "
      (Mercedes-Benz) (Benz)

   (b) an English abbreviation
      " 日本電信電話 （ＮＴＴ） " → " ＮＴＴ "
      (Nippon Telegraph and Telephone)

   (c) the first Katakana character + " 社 "
      " アメリカン・エキスプレス社 " → " ア社 "
      (American Express Corp.)

   (d) the first character of each primitive segment
     "<u>日</u>本<u>航</u>空" → "日航"[1]
     (Japan Airlines)

   (e) some randomly selected characters
     "<u>新</u><u>日</u>本製<u>鉄</u>" → "新日鉄"
     (Shin-nihon Steel)

Locating company name abbreviations is difficult, since many are not identified as companies by either a morphological analyzer or the name recognizer in the preprocessor. Another problem is that the variety of ways of abbreviating names makes it difficult to unify multiple references to one company.

Almost all MUC-5 systems include a string matching mechanism to identify company name abbreviations. These abbreviations are specified in an aliases slot in the company entity object. To the authors' knowledge, none of the systems other than TEXTRACT can detect company name abbreviations of type (d) or (e) above without using a pre-defined abbreviation table.

3. Company name pronouns
Company name pronouns are often used in formal texts. Frequently used expressions include "両社" (*ryosya*: both companies), "同社" (*dosya*: the company), and "自社" (*jisya*: the company itself). As shown in the following examples, resolving the references is particularly important for full understanding of a text. Direct English translation follows the Japanese sentences.

   (a) "X社はY社と提携し、<u>同社</u>の製品を<u>自社</u>ブランドで販売する。"
                       (Y社)        (X社)
     "X Corp. has tied up with Y Corp. and sells products of <u>the company</u> by <u>its own</u> brand name."
                                      (Y Corp.)   (X Corp.)
   (b) "X社はこの分野では最大手。<u>同社</u>の社長は鈴木氏。"
                            (X社)
     "X Corp. is the biggest company in this field. The president of <u>the company</u> is Mr. Suzuki."
                                                  (X Corp.)

Reference resolution for "同社" (*dosya*: the company) is implemented in VENIEX (Doi et al., 1993). VENIEX resolves the pronominal reference in the same way as it identifies missing company references. The CRL/BRANDEIS DIDEROT system presented at MUC-5 simply chooses the nearest company name as the referent of "dosya". This algorithm was later improved by Wakao using corpus-based heuristic knowledge (Wakao, 1994). These systems do not handle pronominalized company names other than "dosya".

The three problems described in this section often cause individual information to be correlated with the wrong company or tie-up-relationship object. To avoid this error, the topic companies must be tracked from the context, since they can be used to determine which company objects an information fragment should be assigned to. Abbreviated and pronominalized company names must be unified as references to the same company.

---

1. "日本" (*nihon*: Japan) and "航空" (*koukuu*: airlines) are the primitive segments in this example.

### 3.3 Discourse segmentation and concept merging

In the joint ventures domain, a tie-up-relationship object contains pointers to other objects such as economic activities (as shown in Figure 1). When a company is involved in multiple tie-ups, merging information into a tie-up relationship according to topic companies sometimes yields incorrect results. Consider the following example:

```
"X Corp. has tied up with Y Corp. X will start selling products
 in Japan next month. Last year X started a similar joint venture
 with Z Inc."
```

Obviously, the sale in the second sentence is related to the tie-up relationship of X and Y. However, since the topic company, which is the subject of a sentence, is X in all three sentences, the sale could also be related to the X and Z tie-up relationship. This incorrect merging can be avoided by separating the text into two blocks: the first two sentences describe the X and Y tie-up, and the last sentence describes the X and Z tie-up. Thus, discourse segmentation is necessary to identify portions of text containing related pieces of information. The CRL/BRANDEIS DIDEROT system segments the joint ventures text into two types of text structures (Cowie et al., 1993). It is not known how well their discourse segmentation performed, however.

Once the text is segmented, concepts or identified pieces of information can be merged within the same discourse segment. For example, the expected income from a joint venture is often stated in a sentence which does not explicitly mention the participating companies; they appear in the previous sentence. In this case, the joint venture concept identifying the companies and the income concept identifying the expected income must be merged so that the latter will be linked to the correct entity objects.

## 4. The solution

This section describes details of TEXTRACT's pattern matcher and discourse processor as well as the system architecture.

### 4.1 TEXTRACT architecture

TEXTRACT is an information extraction system developed for the TIPSTER Japanese domains of corporate joint ventures and microelectronics (Jacobs, 1993; Jacobs et al., 1993). As shown in Figure 2, the TEXTRACT joint ventures system comprises four major components: preprocessor, pattern matcher, discourse processor, and template generator. Because of its shorter development time, the TEXTRACT microelectronics system has a simpler configuration than the joint ventures system. It does not include the template pattern search in the pattern matcher, or the discourse segmentation and concept merging in the discourse processor, as also shown in Figure 2.

In the preprocessor, a Japanese segmentor called MAJESTY segments Japanese text into primitive words tagged with their parts of speech (Kitani, 1991). Next, the name recognizer identifies proper names and monetary, numeric, and temporal expressions. MAJESTY tags proper names which appear in its lexicon; the name recognizer identifies additional proper names by locating name designators such as "社" (*sya*, corresponding to "Inc." or "Corp.")

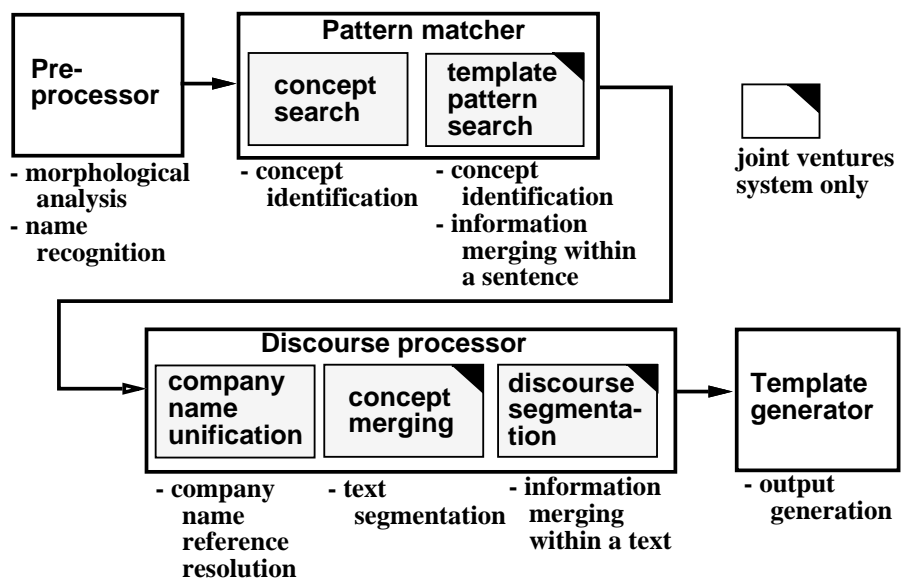

Figure 2: textract system architecture

for company names. The recognizer extends the name string forward and backward from the designator until it meets search stop conditions (Kitani & Mitamura, 1993). The name segments are grouped into units which are meaningful to the pattern matching process (Kitani & Mitamura, 1994). Most strings to be extracted directly from the text are identified by majesty and the name recognizer.

Details of the pattern matcher and discourse processor are given in the following sections. The template generator assembles the extracted information and creates the output described in Section 2.

## 4.2 Pattern matcher

The following subsections describe the *concept search* and the *template pattern search* in the pattern matcher which identify concepts in the sentence. Whereas the former simply searches for key words, the latter searches for phrasal patterns within a sentence. The template pattern search also identifies relationships between matched objects in the defined pattern. In the course of textract development, key words and template patterns were obtained manually by a system developer using a kwic (Key Word In Context) tool and referring to a word frequency list obtained from the corpus.

### 4.2.1 Concept search

Key words representing the same concept are grouped into a list and used to recognize the concept in a sentence. The list is written in a simple format: *(concept-name word1 word2 …)*. For example, key words for recognizing a dissolved joint venture concept can be written in the following way:

```
(DISSOLVED 提携解消 整理 消滅)
              or
(DISSOLVED dissolve terminate cancel).
```

The concept search module recognizes a concept when it locates one of the associated words in a sentence. This simple procedure sometimes yields incorrect concepts. For example, the concept "dissolved" is erroneously identified from an expression such as "*cancel* a hotel reservation". Key-word-based concept search is most successful when processing text in a narrow domain in which words are used with restricted meanings.

The *under-matching* problem occurs when a compound noun in the key word list of a concept fails to match the text because the instance of the compound in the text has been segmented into separate primitive words. To avoid the problem, adjacent nouns in the text are automatically concatenated during the concept search process, generating compound nouns at run-time. The *over-matching* problem, on the other hand, arises when a key word successfully matches *part* of a compound noun which as a whole is not associated with the concept. Over-matching can be prevented by anchoring the beginning and/or end of a key word pattern to word boundaries (with the symbol ">" at the beginning and "<" at the end). For example, "> シリコン <" (*silicon*) must be matched against a single complete word in the text. Since this problem is rare, its solution is not automatic: system developers attach anchors to key words which are likely to over-match.

### 4.2.2 TEMPLATE PATTERN SEARCH

TEXTRACT's pattern matcher is implemented as a finite-state recognizer. This choice of implementation is based on the assumption that a finite-state grammar can efficiently handle many of the inputs that a context-free grammar covers (Pereira, 1990). The pattern matcher is similar to the pattern recognizer used in the MUC-4 FASTUS system developed at SRI (Hobbs et al., 1992).

Patterns for the TEXTRACT template pattern matcher are defined with rules similar to regular expressions. Each pattern definition specifies the concept associated with the pattern. (For the joint ventures domain, TEXTRACT uses eighteen concepts.)

In the matcher, state transitions are driven by segmented words or grouped units from the preprocessor. The matcher identifies all possible patterns of interest in the text that match defined patterns, recognizing the concepts associated with the patterns. For some inputs, the matcher must skip words that are not explicitly defined in the pattern.

Figure 3 shows definitions of equivalent Japanese and English patterns for recognizing the concept *JOINT-VENTURE*. This English pattern is used to capture expressions such as "XYZ Corp. created a joint venture with PQR Inc." The notation "@string" represents a variable matching an arbitrary string. Variables whose names begin with "@CNAME" are called company-name variables and are used where a company name is expected to appear. In the definitions shown, a string matched by "@CNAME_PARTNER_SUBJ" is likely to contain at least one company name referring to a joint venture partner and functioning as the subject in a sentence.

The pattern " は｜が:strict:P" matches the grammatical particles " は " (*wa*) and " が " (*ga*), which serve as subject case markers. The symbol "strict" specifies a full string match (the default in case of template pattern search), whereas "loose" allows a partial string

```
(a)  (JointVenture1 6              (b)  (JointVenture1 3
         @CNAME_PARTNER_SUBJ               @CNAME_PARTNER_SUBJ
         は|が:strict:P                     create::V
         @CNAME_PARTNER_WITH               a joint venture::NP
         と:strict:P                        with::P
         @SKIP                             @CNAME_PARTNER_WITH)
         提携:loose:VN)
```

Figure 3: A matching pattern for (a) Japanese and (b) English

match. Partial string matching is useful for matching a defined pattern to compound words. The verbal nominal pattern "提携: loose:VN" matches compound words such as "企業 提携" (*kigyo-teikei*: a corporate joint venture) as well as "提携" (*teikei*: a joint venture).

The first field in a pattern is the pattern name, which refers to the concept associated with the pattern. The second field is a number indexing a field in the pattern. This field's contents are used to decide quickly whether or not to search within a given string. The matcher only applies the entire pattern to a string when the string contains the text in the indexed field. For efficiency, therefore, this field should contain the least frequent word in the entire pattern (in this case, "提携" (*teikei*) for Japanese and "a joint venture" for English).

The order of noun phrases is relatively unconstrained in a Japanese sentence. Case markers, usually attached to the ends of noun phrases, provide a strong clue for identifying the case role of each phrase (subject, object, etc.). Thus pattern matching driven mainly by case markers recognizes the case roles well without parsing the sentence.

Approximately 150 patterns are used to extract various concepts in the Japanese joint ventures domain. Several patterns usually match a single sentence. Moreover, since patterns are often defined with case markers such as "は" (*wa*), "が" (*ga*), and "と" (*to*), a single pattern can match a sentence in more than one way when several of the same case markers appear in the sentence. The template generator accepts only the best matched pattern, which is chosen by applying the following three heuristic rules in the order shown:

1. select patterns that include the largest number of matched company-name variables containing at least one company name;

2. select patterns that consume the fewest input segments (the shortest string match); and

3. select patterns that include the largest number of variables and defined words.

These heuristic rules were obtained from an examination of matched patterns reported by the system. To obtain more reliable heuristics, a large-scale statistical evaluation must be performed. Heuristics for a similar problem of pattern selection in English are discussed in (Rau & Jacobs, 1991). Their system chooses the pattern which consumes the most input segments (the longest string match), as opposed to TEXTRACT's choice of the shortest string match in its second heuristic rule.[2]

---

2. In Rau and Jacobs' system, the third heuristic rule seems to be applied before the second rule. In this case, there should be little difference in performance between the heuristic rules of the two systems.

Another important feature of the pattern matcher is that rules can be grouped according to their concept. The rule name "JointVenture1" in Figure 3, for example, represents the concept *JOINT-VENTURE*. Using this grouping, the best matched pattern can be selected from matched patterns of a particular concept group instead of from all the matched patterns. This feature enables the discourse and template generation processes to narrow their search for the best information to fill a particular slot.

## 4.3 Discourse processor

The following subsections describe the algorithm of company name reference resolution throughout the discourse. Discourse segmentation and concept merging processes are also discussed.

### 4.3.1 Identifying topic companies

Since no syntactic analysis is performed in TEXTRACT, topic companies are simply identified wherever a subject case marker such as "が" (*ga*), "は" (*wa*), or "も" (*mo*) follows company names. If no topic companies are found in a sentence, the previous sentence's topic companies are inherited (even if the current sentence contains a non-company subject). This is based on the supposition that a sentence which introduces new companies usually mentions them explicitly in its subject.

### 4.3.2 Abbreviation detection and unification

Company name abbreviations have the following observed characteristics:

- MAJESTY tags most abbreviations as "unknown", "company", "person", or "place";

- a company name precedes its abbreviations;

- an abbreviation is composed of two or more characters from the company name, in their original order;

- the characters need not be consecutive within the company name; and

- English word abbreviations must be identical with an English word appearing in the company name.

Thus the following are regarded as abbreviations: "unknown", "company", "person", and "place" segments composed of two or more characters which also appear in company names previously identified in the text. When comparing possible abbreviations against known company names, the length of the longest common subsequence or LCS (Wagner & Fischer, 1974) is computed to determine the maximum number of characters appearing in the same order in both strings.[3]

To unify multiple references to the same company, a unique number is assigned to the source and abbreviated companies. Repeated company names which contain strings appearing earlier in the text are treated as abbreviations (and thus given unique numbers)

---

3. For example, the LCS of "abacbba" and "bcda" is "bca".

1. Step 1: Initialization to assign each entity in $C$ a unique number.

   for $i$ in $C$ do $(1 \leq i \leq cmax)$

      $C[i,\ \text{“}id\text{”}] \leftarrow i$

   done

2. Step 2: Search abbreviations and give unique numbers

   for $i$ in $C$ do $(1 \leq i \leq cmax)$

      if $C[i,\ \text{“}id\text{”}] \neq i$ then           # **already recognized as an abbreviation**

         continue $i$ loop

      $LENSRC \leftarrow$ length of $C[i,\ \text{“}string\text{”}]$

      for $j$ in $C$ do $(i + 1 \leq j \leq cmax)$

         if $C[j,\ \text{“}id\text{”}] \neq j$ then         # **already recognized as an abbreviation**

            continue $j$ loop

         $LEN \leftarrow$ length of $C[j,\ \text{“}string\text{”}]$

         $LCS \leftarrow$ length of the LCS of $C[i,\ \text{“}string\text{”}]$ and $C[j,\ \text{“}string\text{”}]$

         if $LCS \geq 2$ then do

            if $C[i,\ \text{“}eg\text{”}] = \text{“}YES\text{”}$ and $LENSRC = LCS = LEN$ then

               $C[j,\ \text{“}id\text{”}] \leftarrow C[i,\ \text{“}id\text{”}]$    # **an English word abbreviation**

            else if $C[i,\ \text{“}eg\text{”}] = \text{“}NO\text{”}$ and $LCS = LEN$ then     # **an abbreviation**

               $C[j,\ \text{“}id\text{”}] \leftarrow C[i,\ \text{“}id\text{”}]$

         done

      done

   done

Figure 4: Algorithm to unify multiple references to the same company

by the algorithm described in Figure 4. In the pseudocode shown, all identified company names are stored in an associative array named $C$. "Unknown", "company", "person", and "place" segments are also stored in the array as possible abbreviations. Company names are sorted in ascending order of their starting position in the text and numbered from 1 to $cmax$ (Step 1). A company name string which is indexed $i$ can be addressed by $C[i,\ \text{“}string\text{”}]$. A flag $C[i,\ \text{“}eg\text{”}]$ records whether the company name is an English word abbreviation or not.

Step 2 compares each company name in the array $C$ with all names higher in the array (and thus later in the text). When the LCS of a pair of earlier and later company names is equal to the length of the later company name, the later company name is recognized as an abbreviation of the earlier company name. Then, the "id" of the later company name is replaced with that of the earlier company name. The LCS must be two or more characters, and if the abbreviation is an English word, the LCS must be equal to the length of the earlier company name.

At the end of execution, a number is given in $C[i,\ \text{“}id\text{”}]$. If $C[i,\ \text{“}id\text{”}]$ was changed during execution, $C[i,\ \text{“}string\text{”}]$ was recognized as a company name abbreviation.

### 4.3.3 Anaphora resolution of company name pronouns

The approach for reference resolution described in this section is based on heuristics obtained by corpus analysis rather than linguistic theories. Three company name pronouns are the target of reference resolution: "両社" (*ryosya*), "同社" (*dosya*), and "自社" (*jisya*), meaning "both companies", "the company", and "the company itself". They are three of the most frequent company name pronouns appearing in our corpus provided by ARPA for the TIPSTER information extraction project. "Ryosya", "dosya", and "jisya" appeared 456, 277, and 129 times, respectively, in 1100 newspaper articles containing an average of 481 characters per article.

The following heuristics, derived from analysis of pronoun reference in the corpus, were used for reference resolution:

- "ryosya" almost always referred to the "current" tie-up company, with one exception in a hundred occurrences;

- about ninety percent of "dosya" occurrences referred to the topic company when there was only one possible referent in the same sentence, but:

- when more than two companies, including the topic company, preceded "dosya" in the same sentence, about seventy-five percent of the pronoun occurrences referred to the nearest company, not necessarily the topic company; and

- about eighty percent of "jisya" occurrences referred to the topic company.

Two additional heuristic rules were discovered but not implemented in TEXTRACT:

- about four percent of "jisya" occurrences referred to more than one company; and

- about eight percent of "jisya" occurrences referred to entities which are general expressions about a company such as "会社" (*kaisya*: a company).

As a result of the discourse processing described above, every company name, including abbreviations and pronominal references, is given a unique number.

### 4.3.4 Discourse segmentation and concept merging

In the 150 articles of the TIPSTER/MUC-5 joint ventures test set, multiple tie-up relationships appeared in thirty-one articles which included ninety individual tie-up relationships. The two typical discourse models representing the discourse structures of tie-up relationships are shown in Figure 5.

- Type-I: tie-ups are described sequentially
  Descriptions of tie-ups appear sequentially in this model. One tie-up is not mentioned again after a new tie-up has been described.

- Type-II: a main tie-up reappears after other tie-ups are mentioned
  A major difference from the Type-I model is that a description of a main tie-up reappears in the text after other tie-up relationships have been introduced. Non-main tie-ups are usually mentioned briefly.

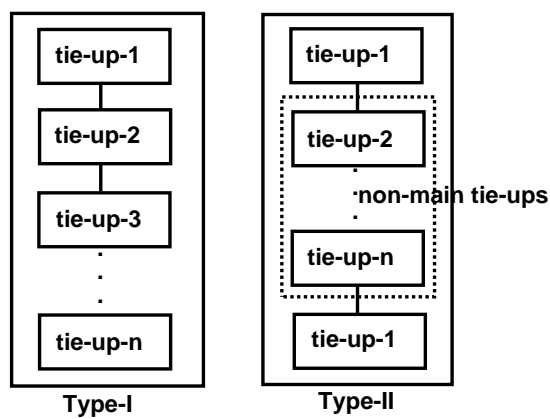

Figure 5: Discourse structure of tie-up relationships

Eleven Type-I structures and thirteen Type-II structures appeared in the thirty-one articles. Seven of the articles contained complicated discourse structures regarding the tie-up relationships.

The two types of text structure described above are similar to the ones implemented in the CRL/BRANDEIS DIDEROT joint ventures system. The difference is only in the Type-II structure: DIDEROT processes all tie-up relationships which reappear in the text, not just the reappearing main tie-up focused on by TEXTRACT.

TEXTRACT's discourse processor divides the text when a different tie-up relationship is identified by the template pattern search. A different tie-up relationship is recognized when the numbers assigned to the joint venture companies are not identical to those appearing in the previous tie-up relationships. DIDEROT segments the discourse if any other related pieces of information such as date and entity location are different between the tie-up relationships. Such strict merging is preferable when the pieces of information in comparison are correctly identified. The merging conditions of discourse segments should be chosen according to the accuracy of identification of the information to be compared.

After the discourse is segmented, identified concepts and extracted words and phrases are merged. Figure 6 shows the merging process for the following text passage which actually appeared in the TIPSTER/MUC-5 test set (a direct English translation follows):

"田辺製薬は8日、西独の医薬メーカー、エー・メルク社の新薬の日本国内での開発、販売をする提携契約を結んだ。新薬の販売ができるようになる5、6年先には、両社が折半出資して合弁会社を設立することも合意した。"

"On the eighth (of this month), Tanabe Pharmaceuticals made a joint venture contract with a German pharmaceutical maker, Merck and Co. Inc., to develop and sell its new medicine in Japan. They also agreed that both companies would invest equally to establish a joint venture company in five or six years when they start selling new medicine."

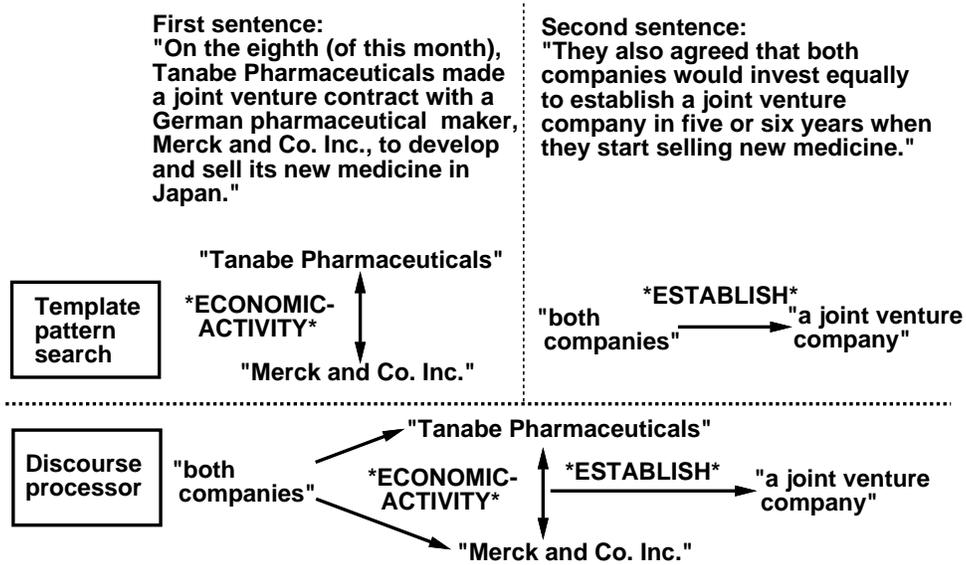

Figure 6: Example of concept merging

The two company names in the first sentence, "田辺製薬" (*tanabe seiyaku*: Tanabe Pharmaceuticals) and "エー・メルク社" (*ei meruku sya*: Merck and Co. Inc.), are identified by either MAJESTY or the name recognizer during preprocessing. Next, the template pattern search locates in the first sentence the "economic activity" pattern shown in Figure 7 (a). The *ECONOMIC-ACTIVITY* concept relating the two companies has now been recognized. The template pattern search also recognizes the *ESTABLISH* concept in the second sentence by the template pattern shown in Figure 7 (b).

After sentence-level processing, discourse processing recognizes that "両社" (*ryosya*: both companies) in the second sentence refers to Tanabe Pharmaceuticals and Merck in the first sentence because they are the current tie-up companies. Since the second sentence does not introduce a new tie-up relationship, both sentences are in the same discourse segment. Concepts separately identified in the two sentences can now be merged because the subjects of the two sentences are the same. The *ESTABLISH* concept is therefore joined to the *ECONOMIC-ACTIVITY* concept.

```
(a)  (EconomicActivityE 6              (b)  (Establish3 6
        @CNAME_PARTNER_SUBJ                    @CNAME_PARTNER_SUBJ
        は | が:strict:P                        は | が:strict:P
        @CNAME_PARTNER_SUBJ                    @CNAME_CREATED_OBJ
        の:strict:P                            を:strict:P
        @SKIP                                  @SKIP
        開発:loose:VN)                          設立:loose:VN)
```

Figure 7: Economic activity pattern (a) and establish pattern (b)

## 5. Performance evaluation

This section shows some evaluation results for TEXTRACT's discourse module. MUC-5 evaluation metrics and overall TEXTRACT performance are also discussed.

### 5.1 Unique identification of company name abbreviations

A hundred joint ventures newspaper articles used for the TIPSTER 18-month evaluation were chosen as a blind test set for this evaluation. The evaluation measures were *recall*, the percentage of correct answers extracted compared to possible answers, and *precision*, the percentage of correct answers extracted compared to actual answers. MAJESTY and the name recognizer identified company names in the evaluation set with recall of seventy-five percent and precision of ninety-five percent when partial matches between expected and recognized strings were allowed, and with recall of sixty-nine percent and precision of eighty-seven percent in an exact matching condition.

Company names that appeared in a form different from their first appearance in an article were considered to be company name abbreviations. Among 318 abbreviations, the recall and precision of abbreviation detection were sixty-seven and eighty-nine percent, respectively. Most importantly, detected abbreviations were unified correctly with the source companies as long as the source companies were identified correctly by MAJESTY and the name recognizer.

The evaluation results clearly show that company name abbreviations were accurately detected and unified with the source companies as long as company names were correctly identified by the preceding processes. It is possible, however, that the simple string matching algorithm currently used could erroneously unify similar company names, which are often seen among family companies.

### 5.2 Anaphora resolution of company name pronouns

The accuracy of reference resolution for "ryosya", "dosya", and "jisya" is shown in Table 1. The numbers in parentheses were obtained by restricting attention to pronouns which referred to companies identified correctly by the preceding processes. Since companies referred to by "ryosya" (both companies) were usually "current" tie-up companies in the joint ventures domain, reference resolution accuracy depended on the accuracy with which tie-up relationships were identified.

| company name pronouns | number of references | resolution accuracy |
|---|---|---|
| " 両社 " (*ryosya*: both companies) | 101 (93) | 64% (70%) |
| " 同社 " (*dosya*: the company) | 100 (90) | 78% (87%) |
| " 自社 " (*jisya*: the company itself ) | 60 (53) | 78% (89%) |

Table 1: Accuracy of reference resolutions

A major cause of incorrect references of "dosya" was the failure to locate topic companies. The simple mechanism of searching for topic companies using case markers did not work well. A typical problem can be seen in the following example: "提携先はＸ社。" (A joint venture partner is X Corp.). Here X Corp is a topic company, but the subject "Ｘ社" (X Corp.) is not followed by a subject case marker. Other errors can be attributed to the fact that "dosya" did not always refer to a topic company as discussed in the heuristic rules of "dosya" reference resolution.

Regarding "jisya" resolutions, five instances which should have referred to multiple companies were bound to a single company. Since multiple companies are usually listed using conjunctions such as "と" (*to*: and) and "、" (comma), they can be identified easily if a simple phrase analysis is performed.

It became clear from this evaluation that resolving "dosya" references to a non-topic company required intensive text understanding. Forty-seven percent of the occurrences of "dosya" and "jisya" were bound to topic companies inherited from a previous sentence. This result strongly supported the importance of keeping track of topic companies throughout the discourse.

## 5.3 Discourse segmentation

Thirty-one of the 150 TIPSTER/MUC-5 evaluation test articles included ninety multiple tie-up relationships. TEXTRACT's discourse processor segmented the thirty-one articles into seventy-one individual tie-up relationship blocks. Only thirty-eight of the blocks were correctly segmented. Main tie-up relationships which reappeared in Type-II discourse structures were not detected well, which caused the structures to be incorrectly recognized as Type-I. This error was caused by the fact that the joint venture relationships were usually mentioned implicitly when they reappeared in the text. For example, a noun phrase, "今回の提携は" (the joint venture this time), which was not detected by the template patterns used, brought the focus back to the main tie-up. As a result, TEXTRACT identified eight percent fewer tie-up relationships than the possible number expected in the TIPSTER/MUC-5 evaluation. This merging error must have affected system performance since the information in the reappearing main tie-up segment would not have been correctly linked to the earlier main tie-up segment.

This preliminary study suggested that recognizing segmentation points in the text should be regarded as crucial for performance. The template pattern matching alone was not good enough to recognize the segmentation points. The discourse processor simply segmented the text when it found a new tie-up relationship. The discourse models, currently unused at run-time in TEXTRACT, could be used to help infer the discourse structure when the system is not sure whether to merge or separate discourse segments. Reference resolution of definite and indefinite noun phrases must also be solved for accurate discourse segmentation in future research.

The accuracy of discourse segmentation might be improved by checking the difference or identity of date and entity location, as well as entity name, when deciding whether or not to merge a tie-up relationship. TEXTRACT did not take date and location objects into account in making segmentation decisions, because TEXTRACT's identification of these objects was not considered reliable enough. For example, the date object was identified with recall of only twenty-seven percent and precision of fifty-nine percent. On the other hand, entities were identified with over eighty percent accuracy in both recall and precision. To avoid incorrect discourse segmentation, therefore, TEXTRACT's merging conditions included only entity names as reliable information.

## 5.4 Overall TEXTRACT performance

250 newspaper articles, 150 about Japanese corporate joint ventures and 100 about Japanese microelectronics, were provided by ARPA for use in the TIPSTER/MUC-5 system evaluation. Six joint ventures systems and five microelectronics systems, including TEXTRACT developed at CMU as an optional system of GE-CMU SHOGUN, were presented in the Japanese system evaluation at MUC-5. A scoring program automatically compared the system output with answer templates created by human analysts. When a human decision was necessary, analysts instructed the scoring program whether the two strings in comparison were completely matched, partially matched, or unmatched. Finally, the scoring program calculated an overall score combined from all the newspaper article scores. Although various evaluation metrics were measured in the evaluation (Chinchor & Sundheim, 1993), only the following error-based and recall-precision-based metrics are discussed in this paper. The basic scoring categories used are: correct (COR), partially correct (PAR), incorrect (INC), missing (MIS), and spurious (SPU), counted as the number of pieces of information in the system output compared to the possible information.

(1) Error-based metrics

- Error per response fill (ERR):

$$\frac{wrong}{total} = \frac{INC + PAR/2 + MIS + SPU}{COR + PAR + INC + MIS + SPU}$$

- Undergeneration (UND):

$$\frac{MIS}{possible} = \frac{MIS}{COR + PAR + INC + MIS}$$

- Overgeneration (OVG):

$$\frac{SPU}{actual} = \frac{SPU}{COR + PAR + INC + SPU}$$

- Substitution (SUB):

$$\frac{INC + PAR/2}{COR + PAR + INC}$$

| domain | ERR | UND | OVG | SUB | REC | PRE | P&R |
|---|---|---|---|---|---|---|---|
| TEXTRACT (JJV) | 50 | 32 | 23 | 12 | 60 | 68 | 63.8 |
| System A (JJV) | 54 | 36 | 27 | 12 | 57 | 64 | 60.1 |
| System B (JJV) | 63 | 51 | 23 | 12 | 42 | 67 | 52.1 |
| TEXTRACT (JME) | 59 | 43 | 28 | 12 | 51 | 63 | 56.4 |
| System A (JME) | 58 | 30 | 38 | 14 | 60 | 53 | 56.3 |
| System B (JME) | 65 | 54 | 24 | 12 | 40 | 66 | 50.4 |

Table 2: Scores of TEXTRACT and two other top-ranking official systems in TIPSTER/MUC-5

(2) Recall-precision-based metrics

- Recall (REC):

$$\frac{COR + PAR/2}{possible}$$

- Precision (PRE):

$$\frac{COR + PAR/2}{actual}$$

- P&R F-measure (P&R):

$$\frac{2 * REC * PRE}{REC + PRE}$$

The error per response fill (ERR) was the official measure of MUC-5 system performance. Secondary evaluation metrics were undergeneration (UND), overgeneration (OVG), and substitution (SUB). The recall, precision, and F-measure metrics were used as unofficial metrics for MUC-5.

Table 2 shows scores of TEXTRACT and two other top-ranking official systems taken from the TIPSTER/MUC-5 system evaluation results.[4] TEXTRACT processed only Japanese domains of corporate joint ventures (JJV) and microelectronics (JME), whereas the two other systems processed both English and Japanese text. TEXTRACT performed as well as the top-ranking systems in the two Japanese domains.

The human performance of four well-trained analysts was reported to be about eighty percent in both recall and precision in the English microelectronics domain (Will, 1993). This is about thirty percent better than the best TIPSTER/MUC-5 systems' performance in P&R F-measure in the same language domain. In the Japanese joint ventures domain, TEXTRACT scored recall of seventy-five percent and precision of eighty-one percent with a core template comprising only essential objects. This result suggests that the current technology could be used to support human extraction work if the task is well-constrained.

---

4. The TEXTRACT scores submitted to MUC-5 were unofficial. It was scored officially after the conference. Table 2 shows TEXTRACT's official scores.

Running on a SUN SPARCstation IPX, TEXTRACT processed a joint ventures article in about sixty seconds and a microelectronics article in about twenty-four seconds on average. The human analysts took about fifteen minutes to complete an English microelectronics template and about sixty minutes for a Japanese joint ventures template (Will, 1993). Thus a human-machine integrated system would be the best solution for fast, high quality, information extraction.

Some TIPSTER/MUC-5 systems processed both Japanese and English domains. These systems generally performed better in the Japanese domains than in the corresponding English domains. One likely reason is that the structure of Japanese articles is fairly standard, particularly in the Japanese joint ventures domain, and can be readily analyzed into the two discourse structure types described in this paper. Another possible reason is a characteristic of writing style: expressions which need to be identified tend to appear in the first few sentences in a form suitable for pattern matching.

The TEXTRACT Japanese microelectronics system copied the preprocessor, the concept search of the pattern matcher, and the company name unification of the discourse processor used in the TEXTRACT Japanese joint ventures system. The microelectronics system was developed in only three weeks by one person who replaced joint ventures concepts and key words with representative microelectronics concepts and key words. The lower performance of the TEXTRACT microelectronics system compared to the joint ventures system is largely due to the short development time. It is also probably due to the less homogeneous discourse structure and writing style of the microelectronics articles.

## 6. Conclusions and future research

This paper has described the importance of discourse processing in three aspects of information extraction: identifying key information throughout the text, i.e. topic companies and company name references in the TIPSTER/MUC-5 domains; segmenting the text to select relevant portions of interest; and merging concepts identified by the sentence level processing. The basic performance of the system depends on the preprocessor, however, since many pieces of identified information are put directly into slots or are otherwise used to fill slots during later processing. TEXTRACT's pattern matcher solves the matching problem caused by the segmentation ambiguities often found in Japanese compound words. The pattern matching system based on a finite-state automaton is simple and runs fast. These factors are essential for rapid system development and performance improvement.

To improve system performance with the pattern matching architecture, an increase in the number of patterns is unavoidable. Since matching a large number of patterns is a lengthy process, an efficient pattern matcher is required to shorten the running time. Tomita's new generalized LR parser, known to be one of the fastest parsers for practical purposes, skips unnecessary words during parsing (Bates & Lavie, 1991). The parser is under evaluation to investigate if it is appropriate for information extraction from Japanese text (Eriguchi & Kitani, 1993). Pattern matching alone, however, will not be able to improve the system performance to human levels in a complicated information extraction task such as TIPSTER/MUC-5, even if the task is well-defined and suitable for pattern matching. More efforts should be made in discourse processing such as discourse segmentation and reference resolution for definite and indefinite noun phrases.

The research discussed in this paper is based on an application-oriented, domain-specific, and language-specific approach relying on patterns and heuristic rules collected from a particular corpus. It is obvious that the patterns and heuristic rules described in this paper do not cover a wide range of applications, domains, or languages. The empirical approach described here is worth investigating even for an entirely new task, however, since it can achieve a high level of system performance in a relatively short development time. While linguistic theory-based systems tend to become complex and difficult to maintain, especially if they incorporate full text parsing, the simplicity of an empirically-based, pattern-oriented system such as TEXTRACT keeps the development time short and the evaluation cycle quick.

Corpus analysis is a key element in this corpus-based paradigm. It is estimated that corpus analysis took about half of the development time for TEXTRACT. Statistically-based corpus analysis tools are necessary to obtain better performance in a shorter development time. Such tools could help developers not only extract important patterns and heuristic rules from the corpus, but also monitor the system performance during the evaluation-improvement cycle.

## Acknowledgements

The authors wish to express their appreciation to Jaime Carbonell, who provided the opportunity to pursue this research at the Center for Machine Translation, Carnegie Mellon University. Thanks are also due to Teruko Mitamura and Michael Mauldin for their many helpful suggestions.